\documentclass{article}
\usepackage{graphicx} 

\title{Language Models as Semiotic Machines: Reconceptualizing AI Language Systems through Structuralist and Post-Structuralist Theories of Language}
\author{Elad Vromen}
\date{October 2024}

\begin{document}

\maketitle
\begin{abstract}
This paper proposes a novel framework for understanding large language models (LLMs) by reconceptualizing them as semiotic machines rather than as imitations of human cognition. Drawing from structuralist and post-structuralist theories of language—specifically the works of Ferdinand de Saussure and Jacques Derrida—I argue that LLMs should be understood as models of language itself, aligning with Derrida's concept of ''writing'' (l'écriture).

The paper is structured into three parts. First, I lay the theoretical groundwork by explaining how the word2vec embedding algorithm operates within Saussure’s framework of language as a relational system of signs. Second, I apply Derrida's critique of Saussure to position ''writing'' as the object modeled by LLMs, offering a view of the machine's ''mind'' as a statistical approximation of sign behavior. Finally, the third section addresses how modern LLMs reflect post-structuralist notions of unfixed meaning, arguing that the ''next token generation'' mechanism effectively captures the dynamic nature of meaning.

By reconceptualizing LLMs as semiotic machines rather than cognitive models, this framework provides an alternative lens through which to assess the strengths and limitations of LLMs, offering new avenues for future research.
\end{abstract}
\newpage
\section*{Introduction}

Recent advancements in artificial intelligence, particularly in language models, have sparked lively debates in both casual conversations and academic circles. Some dismiss these models as mere poor imitations\footnote{Noam Chomsky, ``Noam Chomsky: The False Promise of ChatGPT'', New York Times, 2023} of the human mind, while others herald them as ``innovative scientific explanations that push aside all previous linguistic theories''\footnote{Steven T. Piantadosi, ``Modern language models refute Chomsky’s approach to language'', 2023}. At the heart of this debate lies a fundamental question: how similar—or different—are these AI systems to the way humans acquire and use language? And what, if any, is their contribution to our understanding of language?

Two long-standing issues often emerge in these discussions:
\begin{enumerate}
    \item The vast gap between the training data used to develop language models and the limited linguistic input that children are exposed to.
    \item The contrast between our human notion of "understanding" the meaning of a statement and AI’s method of generating text through statistical prediction of the next word.
\end{enumerate}

While these concerns are valid, I believe the confusion stems from a fundamental lack of clarity: what exactly are language models modeling? Instead of viewing these systems as imitations of human minds or thinking processes (as the term ``artificial intelligence'' might imply), I propose that we see them as \textit{semiotic machines}—representations of language itself as a system of signs, specifically in the form of writing.

By adopting this perspective, we can better grasp the true nature of language models, enhance their utility, and more accurately assess their scientific and epistemic contributions. I aim to apply structuralist and post-structuralist theories to analyze the semiotic structure and concept of meaning in language models. This framework will help us both understand the epistemic status of these models and consider them as potential empirical evidence for broader philosophical theories of language.

To do this, I will first establish the theoretical groundwork by tracing the development of artificial intelligence algorithms as opposed to the formal encoding of letters in computers, drawing from Ferdinand de Saussure's structuralist linguistic theory in \textit{Course in General Linguistics}\footnote{Saussure, F. de. (1959). \textit{Course in General Linguistics}. (W. Baskin, Trans.). New York: Philosophical Library.}.

Next, I will clarify the object of modeling in language models through Jacques Derrida’s concept of \textit{l’écriture} (writing) as presented in \textit{Of Grammatology}\footnote{Derrida, J. (1976). \textit{Of Grammatology}. (G. C. Spivak, Trans.). Baltimore: Johns Hopkins University Press.}.

Finally, I will connect the evolution of language models—from primitive systems to the advanced models we have today—with the shift from structuralist to post-structuralist theories of meaning, focusing on Derrida's \textit{Signature Event Context}\footnote{Jacques Derrida, ``Signature Event Context'', \textit{Limited Inc}, Northwestern University Press, 1988}.
\newpage 

\section*{Part One - Structuralism and Word Embedding: First Steps Toward a Semantic Theory of AI}

When selecting a theory of language—especially concerning semantic meaning in Large Language Models (LLMs)—it is crucial to consider specific constraints that ensure the theory is both adequate and well-reasoned.

The first constraint is that the theory should be non-referential—it should not rely on the traditional distinction between sign and reference or define semantic meaning by denotation to objects in the world. This is because language models function entirely within a closed textual system, without engaging with any external world (I will not discuss RLHF techniques in this article, which requires further consideration, and will primarily focus on LLM pretraining).

The second constraint is that the theory should ideally be non-mentalistic. In other words, it should avoid requiring concepts like `'consciousness'' or ``mind'' as we understand them in humans. While it’s possible to develop a mentalistic theory, this would involve proving that neural networks can achieve some form of consciousness—a challenge that may be better avoided.

\subsection*{Signifier, signified: what is missing in binary representation of words}

The fundamental challenge in Natural Language Processing (NLP) arises from the gap between the computer's ability to handle representations as a computational machine and language as a system of meaning. In other words, how can binary sequences in a computer’s memory represent the semantic meaning of words and sentences?

Ferdinand de Saussure's work\footnote{Saussure, 1959} offers insight here. He posits that language is a system of signs. In his view, each sign consists of two parts: the signifier (the form, such as a word or sound) and the signified (the concept or meaning it represents). Crucially, Saussure argues that the relationship between these two components is arbitrary (there's no inherent reason why the word "t-r-e-e" represents the concept of a tall, woody plant, for example). Consequently, a deterministic representation of the signifier through other signifiers (such as representing a letter by a number, like in the ASCII table) serves only to represent the signifier, not the signified. This deterministic mapping allows computers to store textual information, which can later be decoded by a human mind that knows the language. However, this process doesn't help machines decode the deeper, signified layer of meaning on their own.

However, Saussure’s theory opens a promising path for modeling meaning. This is because his conception of language is not subjective. Language, according to Saussure, is not merely a cognitive function of individuals, but a collective product—a socio-linguistic network that fixes signifiers and their relationships, forming a meaningful structure. A human mind can engage in language, but language's true essence resides in the intersubjective space. Furthermore, Saussure provides the first step in addressing the modeling problem: the structuralist conception of meaning. In this framework, the meaning of a sign arises from its relative position within the overall system, rather than from any inherent value. This idea of semantic value provides the non-referential foundation we are looking for.

\subsection*{Word2Vec}

Word2Vec (w2v) was the first algorithm to significantly enhance language models' capability to represent semantic information. It tackles the 'language problem' by generating a vector representation (an array of numbers) for each word, based on its relationship to other words in the dataset. The algorithm, a neural network, takes the current word as input and predicts the next word in the sequence. Through this iterative process, the network learns to refine its predictions by statistically evaluating the contexts in which words appear within sentences. Essentially, the algorithm creates a multi-dimensional topographic map of the statistical likelihood of word substitution.

The output of Word2Vec is a vector representation for each word, which encodes not only the deterministic signifier but also the contextual meaning of the word within the dataset. I contend that this contextual representation constitutes the signified aspect of these signs, within a conceptual system, thus imbuing the vector array produced by Word2Vec with the qualities of a Saussurean sign.

\subsection*{Large Language Models}

Large language models are a form of artificial intelligence that have achieved near-human proficiency in language use at a communicative level. Several technical aspects relevant to language and meaning arise from these models.

First, it’s important to note that the context-based representation introduced by Word2Vec remains foundational in training large language models. The GPT series exemplifies this technology, and the paper introducing GPT-2\footnote{Alec Radford et al., ``Improving Language Understanding by Generative Pre-Training'', 2018} emphasized generating contextual language representations before adapting to specific tasks. In other words, these models focus on capturing the relationships between linguistic elements in the dataset as a precursor to solving particular tasks. Understanding Word2Vec's semantic topography offers an intuitive explanation of this pre-training process.

Second, It is essential to recognize that large language models go beyond the basic Word2Vec framework. The representations created by these models encompass more than just individual words; they include sentences and other linguistic structures (A thorough technical explanation of these complexities would involve discussions of RNNs and LSTMs, but this falls outside the scope of this article).

Third, what sets large language models apart from other models is the number of parameters (i.e., the complexity of their modeling functions) and the size of the datasets used for training. Current language models are exposed to an extensive amount of information—representing significant portions of all content available on the internet—which causes the object modeled by these language models to approximate language itself.

To summarize the first part, we have observed that modeling the semantic meaning of text involves algorithms that create contextual representations of signs, aligning with Saussure’s structuralist perspective. In the next section, we will explore the textual object that these algorithms model and explain why its statistical analysis is a sufficient approach for capturing semantic meaning.
\newpage 
\section*{Part Two - Data as "Writing"}

An important observation is that every language model develops its representation of the signified and, consequently, its ability to use language through exposure to a multiplicity of written signifiers. Therefore, I will rely on Jacques Derrida's deconstruction of the pair of concepts: writing and speech. It is significant that in the introduction to \textit{Of Grammatology}, Derrida states that ``the entire field covered by the cybernetic program will be the field of writing... To suppose that the theory of cybernetics can dislodge by itself all the metaphysical concepts— all the way to the concepts of soul, of life, of value, of choice, of memory— which until recently served to separate the machine from man it must conserve, until its own historico- metaphysical belonging is also denounced, the notion of writing''.\footnote{Derrida, 1976, 9}

Derrida's reading of Saussure reveals a distinct preference in Saussure’s work for phonetic pronunciation, driven by the belief that it is ``closer'' to the signified or the ideal mental content. This contrasts with writing, which is depicted as a ``sign of a sign'', representing a system external to the primary coupling of the phonetic signifier to the signified. Derrida identifies this hierarchy as part of logocentrism and the ancient perception that equates meaning with the speaker's intention. Moreover, logocentrism preserves "the distinction between the sensible and the intelligible''\footnote{Derrida, 1976, 13} and ``the reference to a signified able to ''take place'', in its intelligibility, before its ''fall'', before any expulsion into the exteriority of the sensible here below''\footnote{Derrida, 1976, 14}

This assumption is fundamentally challenged in the philosophy of language within artificial intelligence. When we discuss the linguistic qualities of a language model, we always start with written signs. Any ascent to the realm of meaning is a byproduct of extensive statistical analysis of an array of exclusively written signs. The phonetic medium does not exist, and there is no basis for assuming its precedence or proximity to a ‘'mind’' that serves as the seat of the signified. The ''meaning'' of language models results from mapping relationships between instances of signs. The 'intelligible' emerges from the ''sensible'', empirically.

The perception of writing's inferiority to speech that Derrida identifies in Saussure arises from a set of hierarchical dichotomies embedded in logocentrism.

\subsection*{Presence/Representation, Reality/Image}

In logocentric thought, ''one submits the sign to the question of essence''\footnote{Derrida, 1976, 20}. There is a precedence of presence over the sign, which is perceived as representation. One can summarize the logocentric perception as follows: the world is a collection of states of affairs. Thought is an image of the world, and language is an image of thought. The truth values of all propositions are measured concerning the layer preceding them, with the world serving as the denoted reference point.

Derrida proposes to liberate the sign from its subordination to reality. This allows ``
Reading, and therefore writing, the text...(to be) “originary” operations with a view to a sense that they do not first have to transcribe or discover, which would not therefore be a truth signified in the original element and presence''.\footnote{Derrida, 1976, 20} The representational model is thus turned on its head. Instead of writing being a parasite on language, thought, and the world, the system of signs is established as primary, and writing becomes the paradigm of an independent system of signs.

Let us understand this shift as a foundation for the algorithmic model of meaning. A language model is a statistical approximation of the relationships between signifiers, leading to a contextual representation. This statistical ''projection'' is the only notion of signifieds that can be attributed to the computational system. In fact, rather than ''thought'' being a projection of a ''world'', and language a projection of thought, the language model represents a ''thought'' or meaning that is a projection of writing. Writing consists of a collection of unique signifiers that appear in specific sequences. Post-textual thought (artificial intelligence) serves as a statistical description of the relationships between occurrences of signifiers in the dataset.

\subsection*{Language-Thought-World Models}

We can try to describe an extension of the logocentric model, which will include language models, in the following way:
\begin{figure}[h]
    \centering
    \includegraphics[width=0.8\textwidth]{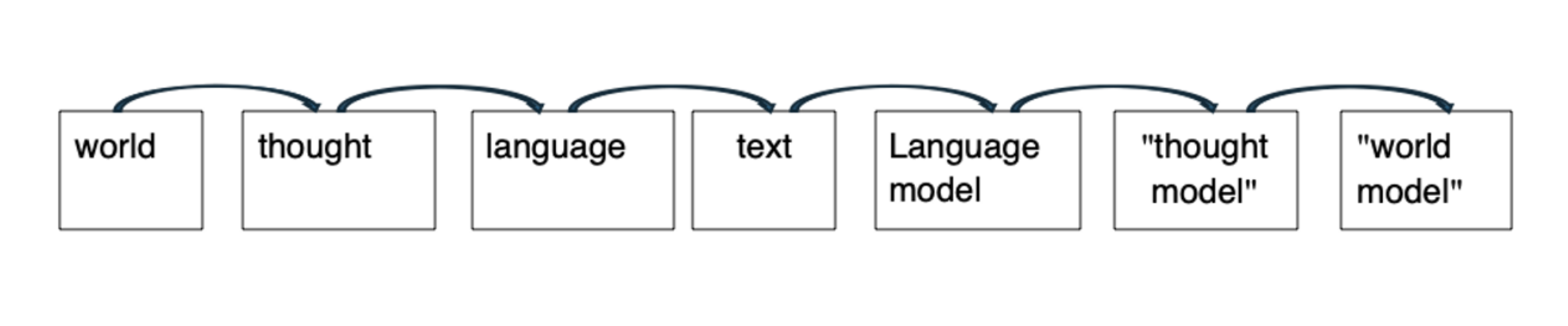}
    \caption{Extension of the logocentric model including language models}
    \label{fig:language_model}
\end{figure}

When we understand the operation of language in this manner, we create a symmetrical analogy between the real world at one end and its representation by a virtual world that unfolds as a projection of the modeled writing at the other end. Such a model presents the problem of language modeling as a question of degrees of representation.
\newpage
Let's refine the diagram using the Saussurian model, as described by Derrida:

\begin{figure}[h]
    \centering
    \includegraphics[width=0.8\textwidth]{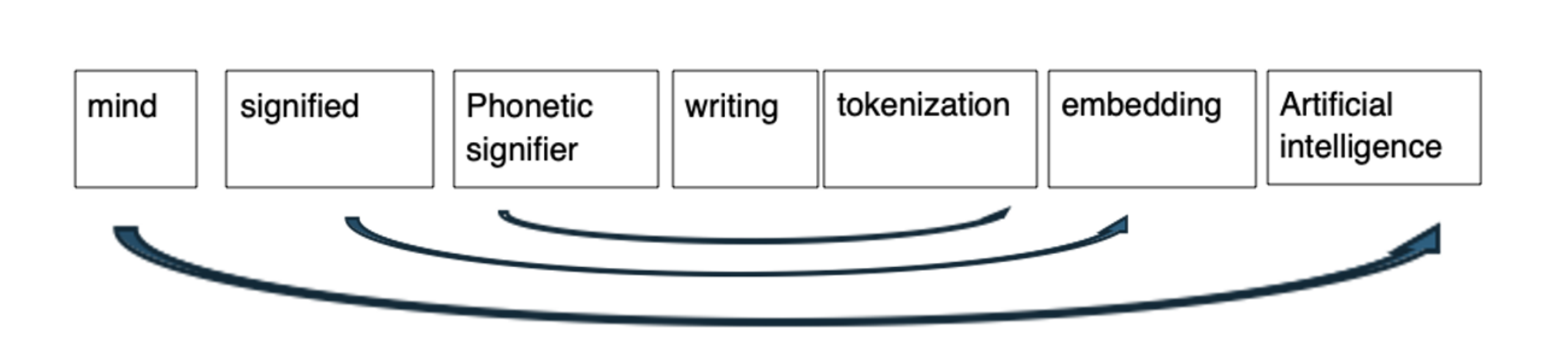}
    \caption{refined logocentric model}
    \label{fig:language_model}
\end{figure}

The diagram draws a parallel between the phonetic signifier in traditional linguistics and the token inserted into the algorithm in AI language models. Both are signifiers—they are nominal yet arbitrary. Their meanings are determined by their connections to signifieds. The coupling between the phonetic signifier and the ideal signified in Saussure corresponds to the relationship between tokenization and the embedding as a coordinate within the network model in which it occupies a position in N-dimensional space. The resulting network model serves as the modeling of the system of signifieds.

Both diagrams position writing as an intermediate stage. On the side of ''presence'' (and mind), writing is a parasitic system, a ''sign of a sign'', in contrast to the phonetic signifier’s coupling to the signified.

Conversely, from the ''perspective'' of artificial intelligence, writing is the presence in relation to the ''thought model'' and the ''world model'', and it constitutes the source that the signified network model represents. From this perspective, writing functions as a world.

\subsection*{Inside/Outside: Artificial Intelligence as the Ultimate Sin}

Derrida identifies Saussure's and logocentrism's perception of writing as an external to the essential connection between the phonetic signifier and the mental signified. ``Writing would thus have the exteriority that one attributes to utensils; in addition it is an imperfect tool and a dangerous, one would say almost maleficent, technique... a sin... inversion of the natural relationship between the soul and the body...''\footnote{Derrida, 1976, 37}

Derrida's originality allows him to uncover the cultural bias against writing and to highlight the radical positions he describes. In contrast, the attitude toward artificial intelligence is far more pronounced. Artificial intelligence is viewed as the ultimate ''sin''. The machine's use of language is inherently detached from the unity of language in the mind. According to the logocentric model of thought, writing is peripheral, while machine thought is even further removed—a derivative of writing itself (emerging from the statistical analysis of text).

Derrida exposes the logocentric fallacy: the longing for a clear hierarchy between origin and copy, reality and representation. Saussure and his tradition always strive to restore the natural order, resisting ''temptation'' and ''artificiality''. Consequently, signs of inverted order are perceived as perversions. However, this hierarchical perception is misguided. Writing is woven into the essence of language. Derrida bases his critique on a commitment to Saussure's own principle of the arbitrariness of the sign- ''The thesis of the arbitrariness of the sign should interdict a radical distinction between the linguistic and the graphic sign''\footnote{Derrida, 1976, 48}

For Derrida, writing is the fundamental paradigm of language. Language consists of a collection of signifiers, whether vocal or written, governed by ``the regulated play of their differences.''\footnote{Derrida, 1976, 48}
 This differentiated system generates 'meaning' independently of the signifier's proximity to any mind or its representation of any world. The sign occupies a specific position within the system as a difference, distinct from any ''background''. No naturalistic entity serves as a source that language represents or from which meaning is derived. Instead, meaning is continuously established through differentiation. Hence, Derrida's concept of meaning as an attribute of the system of signs leads to a non-mentalistic semantic theory. This approach weaves both ''writing'' and language models from peripheral representations into the core and essence of what semantic meaning truly is.
\newpage 
\section*{Part Three - Post-Structuralism and Text Generation}

Up to this point, we have established a non-referential and non-mentalistic semantic theory that can be attributed to both humans and artificial intelligence. However, we are left with a structuralist conception of meaning using the analogy of word embedding—as if the meaning of a sign (the signified) is a nominal value which, although it is a function of the sign's position in the system of signs, remains fixed. This concept has two significant issues:
\begin{enumerate}
    \item From the philosophy of language perspective, it is evident that this is not how meaning operates. ``The value of the notion of literal meaning appears more problematic than ever.''\footnote{Derrida, ``Signature Event Context'', 2}
    \item This perception reduces large language models to mere embedding creation algorithms. In practice, however, modern large language models function more dynamically, primarily performing the task of text generation.
\end{enumerate}

Fortunately, we can address these two developments by referencing "Signature Event Context" and transitioning from a structuralist to a post-structuralist theory. The discussion will include a brief technical overview of the mechanism of text generation and a theoretical shift from the term ``sign'' to the term ``signifying form''.

\subsection*{Theoretical Position}

In the article ``'Signature Event Context'... in, well, context''\footnote{Joshua Kates, ``Signature Event Context''... in, well, context, Journal of the Philosophy of History, 12 (2018)}, the genealogy of linguistic-philosophical approaches to the origin and locus of semantic meaning is explored through the contrast between language (\textit{langue}) and discourse (based on Saussure's \textit{parole}). The article describes, on one hand, the structuralist school, which locates semantic meaning in language: ``language, accordingly, precedes discourse and makes it possible by providing a store of meanings.''\footnote{Kates, ``Signature Event Context''... in, well, context, p. 119} On the other hand, it discusses representatives of the ``second linguistic turn'', who argue that semantic meaning should be situated precisely in discourse. It also notes that proponents of the second linguistic turn were followers of phenomenology, and from their perspective, discourse (and the semantic meaning they attributed to it) was inherently tied to the need for reference.

A third theoretical tradition worth mentioning is the analytic tradition, with John Searle as its (self-appointed) representative in the discussion on SEC. Searle introduces the distinction between ''type''—the abstract category of the sign, which he argues inherently contains its potential semantic meaning—and ''token''—the specific instance of a sign in discourse, whose semantic value is determined by the possibilities inherent in its type.

In contrast, Derrida represents a post-structuralist approach that opposes the second linguistic turn due to his principled opposition to the ``metaphysics of presence''. His goal is to create a semantical framework that is not based on the transcendental signified (as we saw in \textit{Of Grammatology}), does not ''collapse'' back into dependence on reference and presence, and avoids the complete subordination of the token to the type. To achieve this, he seeks to establish a new concept of signs that is not conditioned by the transcendental signified, an \textit{a priori} ideal linguistic entity (the type), or by presence.

\subsection*{Signifying Form}

Derrida's concept of writing views writing and language as functional systems that are shaped by the sum of sign occurrences within them. Writing has two necessary and distinct formal characteristics:
\begin{enumerate}
    \item \textbf{Writing as a Collection of Sign-Occurrences:} Writing consists of signifying forms—specific instances of signs—to which additional occurrences are continuously added. These sign occurrences are not simply the 'theoretical bag of words' found in language, but actual instances of these signs in writing (and in language more broadly, where writing serves as the paradigm). They represent ``single instances of language's employment''.\footnote{Kates, ``Signature Event Context''... in, well, context, p. 133} Two occurrences of the same word in different contexts are distinct signifying forms. However, these sign occurrences can preserve meaning in a non-spatial and non-temporal manner, persisting materially over time. This property makes iterability (repeatability) a necessary feature of the system.
    \item \textbf{Iterability and the Rupture of Presence:} Iterability gives writing its nature as a 'code' that inherently involves ``a rupture in presence''.
    \begin{enumerate}
        \item \textbf{The Structural Orphaning of Writing from the Addressee:} ``Communication must be repeatable—iterable—in the absolute absence of the receiver or any empirically determinable collectivity of receivers.''\footnote{Derrida, ``Signature Event Context'', p. 7}
        \item \textbf{The Structural Orphaning of Writing from the Addressor:} ``To write is to produce... a machine... which my future disappearance will not hinder from functioning.''\footnote{Derrida, ``Signature Event Context'', p. 9}
        \item \textbf{The Limitation of the Original Context's Power to Fix Meaning:} ``The limiting of the concept of context... in as much as its rigorous theoretical determination... is rendered impossible by writing.''\footnote{Kates, ``Signature Event Context''... in, well, context, p. 133}
    \end{enumerate}
\end{enumerate}

The ''signifying forms'' are the discursive moments of token occurrences: ``They include a moment of discourse, of language in use... tokens.''

According to Derrida, these forms, these instances of discourse, are not in opposition to language (in terms of the discourse-language separation), and are not hierarchically subordinate to an \textit{a priori} language, but part of the unified phenomenon of language, which is writing. He cancels the hierarchy between the \textit{a priori} 'language' (\textit{langue}) and its use in discourse, and establishes writing as a formal 'signification' space that constitutes the space in which semantic meaning is actively created by the fluidity of multiple unfixed contexts.

Thus, a more complex concept is created than any of the previous concepts separately, containing an infinite process of dialectical preservation and change. The differentiation process itself occurs in writing incessantly, but not in a ''one-time'' manner of an event, but as a timeless, spaceless process, of which iterability is an integral part. In this, the concept of meaning derived from the concept of writing escapes the ''sources'' that fix it:

On one hand, the signifying forms in writing are not subject (in terms of their semantic value) to an \textit{a priori} set of rules of language or TYPE, but are open to change and expansion.

On the other hand, signifying forms are not confined to their singular ''occurrence'' in discourse. Instead, they persist and remain open to the indeterminacy created by iterability—the indeterminacy resulting from the removal of the sender, recipient, reference, and context:

``A written sign... does not exhaust itself in the moment... (a) force of rupture... separating... from... the internal contextual chain... (and) also from all forms of present reference.''\footnote{Derrida, ``Signature Event Context'', p. 9}

This dialectical process is explored in ``Signature Event Context'' in relation to Husserl's concept of ideality. Ideality refers to the element in language that is preserved over time, allowing concepts to be understood in a timeless and spaceless way, enabling shared understanding across different individuals.

``Ideality applies to what can and must be referred to as the same, not similar, across various temporal instantiations.''\footnote{Kates, ``Signature Event Context''... in, well, context, p. 134}

However, ideality, by enabling the timeless and spaceless retention of meaning, also facilitates iterability. Iterability, in turn, allows for the involvement of new speakers, recipients, and contexts, thereby preventing the final stabilization of meaning. While there is always an ''intention'' in language use, meaning is not fixed to a single instance but is shaped by a ''matrix of repetition''—the range of different contexts in which the sentence can appear at another time, in another place, with different senders and recipients, all stemming from iterability.

``meaning... ultimately prove enmeshed in a greater matrix of repetition''\footnote{Kates, ``Signature Event Context''... in, well, context, p. 137}

Therefore, we can never view ideality—the timeless and spaceless nature of meaning—as a nominal representation. It always remains open due to the possibility of iterability. Iterability expands the notion of context beyond traditional views, such as Frege's context principle (which ties the meaning of a statement to the sentence it appears in) or de Saussure's focus on the synchronic system of language. Instead, it suggests that the meaning of a statement is shaped by all possible contexts—past, present, and crucially, future. This inclusion of potential future contexts introduces inherent indeterminacy to meaning. Context is no longer confined to what has been or what is, but also to what might be, rendering the meaning of any statement perpetually open and unfixed.

The multiplicity of contexts gives the signifier a new form as a signifying form. The signifying form is not a container of a single, fixed meaning tied to an exclusive signified; rather, it operates across different contexts and allows meaning to evolve through iterative writing.

``there are only contexts without any center or absolute anchoring''\footnote{Derrida, ``Signature Event Context'', p. 12}

\subsection*{Text Generation}

In simplified terms, the operation of language models can be described as follows:

Text generation occurs after the model has been trained and developed a system of representations from the dataset. In other words, the model's weights have been calibrated, having learned both semantic and syntactic information about language (this is the concept of pre-training).

A trained base language model, such as GPT, performs one core task: it receives a sequence of signs (tokens, which may represent fractions of words) as input and outputs a single token at a time, based on the conditional probability of that token relative to the input. For clarity, let's call this task a generation unit. Thus, when a language model generates a 10-word sentence, its output consists of 10 generation units, where each subsequent unit includes the output of previous units as part of its input.

A crucial aspect of understanding text generation is that the selection of tokens in each generation unit involves a sampling function. The representations learned during training span a vector space, where the current input is mapped. Through the attention mechanism, the model dynamically prioritizes different parts of the input with varying importance. This mechanism allows the model to weigh the relevance of each input token when predicting the next token, enabling it to capture long-range dependencies and context-specific meanings, and thus locate the most relevant preceding tokens. The spatial view highlights that the next token is not a fixed option but part of a spectrum of possibilities. In fact, the value of each generation unit is not fixed.

Firstly, even minor changes in the model's input (the prompt) can significantly alter the probability distribution of words (via the attention mechanism). For example, using different verbs (explain/describe/justify) can lead to substantial variations in the output.

Secondly, research on active language models (those optimized for conversation with humans) found that selecting the single most probable word (token) deterministically results in generic and often 'boring' output. Moreover, determinism impairs performance in most language tasks. Complex language use requires more than merely predicting the most likely next word—it demands an element of what could be seen as ''creativity'' from a psychological perspective, paradigmatic selection from a linguistic standpoint, or statistical noise from a technical perspective.

This is where the temperature mechanism comes into play. Temperature is a parameter that controls the model's level of ''creativity'' during text generation. It regulates how much the model can deviate from the most likely word and how much randomness is introduced in the token selection process.

An important consideration regarding these random characteristics is that language models generate sequences of consecutive generation units, and therefore, changes have a cumulative effect. For example, if the first word generated in two instances differs (let's call them w1 and w2), the input for the next generation unit will be [input + w1] in the first instance and [input + w2] in the second. Consequently, the cumulative effect of the changes compounds throughout the response, amplifying the differences.

Thus, we can summarize the text generation process as follows. The language model learns a probabilistic function of word occurrences across the training dataset—a distribution of signifying forms in various contexts. It constructs a topographic model of language and, when given a prompt (a ''piece of language''), spatially locates it within the vector space and matches it with a token that is probabilistically appropriate. Rather than selecting a single possibility, the model operates within a range of possibilities, and the selection method is not deterministic but one of sampling.

\subsection*{Meaning as a Distribution Over Multiple Contexts in the Dataset}

As we've seen through the analysis of \textit{Signature Event Context} (SEC), the structuralist perspective, as represented by the embedding analogy, falls short. It maintains a conception of meaning as fixed and nominal. In contrast, the post-structuralist idea of the contextual permeability of meaning offers a more fitting explanation. This notion better aligns with modern language models' dynamic and flexible operation, presenting a more comprehensive understanding of how semantic meaning is operationalized within them. 

I propose a new phrasing for understanding the post-structuralist concept of meaning. Meaning is a distribution over a space of sign occurrences (signifying forms), defined by a (non-final) metric of repetitions across different contexts in writing or language. This definition aligns with the mechanisms we've described. The language model is exposed to the entirety of signifying forms within the dataset and builds a contextual representation of the semantic space linked to each signifying form, shaped by the occurrences it encounters. Meaning here is not a nominal value but a distribution function—a mechanism that associates the probable occurrence of the next word, based on the specific context through the attention mechanism, using a statistical approximation of all prior occurrences of the signifying forms across various contexts in the data.

The sensitivity of language models to prompts is a fundamental aspect of language. Although the contextual representation is established during pre-training, the prompt, combined with the attention mechanism, allows the semantic meaning of a signifying form to remain open to new contexts, continually reshaping its semantic value. This contextual positioning occurs within a subspace of the language distribution but is undetermined, and its final meaning remains contingent on its usage. Additionally, the ability of large language models (LLMs) to provide different responses to the same question is crucial. This reflects the inherently open and unfixed nature of the semantic meaning of a statement, resulting from the separation of writing from senders, receivers, and fixed contexts, which makes it impossible to fully predetermine the semantic value of certain statements.
\newpage 
\section*{Summary}

When we understand the object of modeling of a language model as language itself—specifically writing—and not as the human mind, we realize that the two problems presented earlier are not actual problems.

Regarding the comparison between children's learning and language models, a language model operates by modeling the occurrences of signs in writing as a relational system. In this system, the value of each sign arises from its relationships with all other signs. When a language model is trained on a smaller dataset, the occurrences of signs and their interrelationships change, resulting in a model of a language that differs from our own. Consequently, this limitation renders it insufficient for our communicative needs (as discussed in Parts 1 and 2).

Regarding the gap between understanding and the statistical prediction of the next word, we demonstrated that meaning is not a nominal value and that a distribution over different contexts aligns with a post-structuralist conception of language. Therefore, using a sampling procedure from this distribution is an appropriate way to describe the behavior of meaning in language (as discussed in Part 3), independent of the understanding process employed by humans.

We have established that semantic meaning is a product of writing and that a computational system that approximates the behavior of signs in writing aligns with the post-structuralist notion of meaning. However, we are left to question the boundaries of this modeled semantic meaning concerning concepts such as ``knowledge'' and ``truth'', which lie beyond the scope of deeper exploration in this paper.

\end{document}